\documentclass{article}

\usepackage[preprint]{neurips_2019}
\usepackage[utf8]{inputenc} 
\usepackage[T1]{fontenc}    
\usepackage{hyperref}       
\usepackage{url}            
\usepackage{booktabs}       
\usepackage{amsfonts}       
\usepackage{nicefrac}       
\usepackage{microtype}      

\usepackage{graphicx}
\usepackage[ruled,vlined]{algorithm2e}
\usepackage{booktabs}
\usepackage{amsmath}

\usepackage{natbib}
\setcitestyle{authoryear,open={(},close={)}}
\hypersetup{colorlinks,linkcolor={blue},citecolor={blue},urlcolor={red}}

\title{AA3DNet: Attention Augmented Real Time 3D Object Detection}

\author{%
  Abhinav Sagar\\
  Vellore Institute of Technology\\
  Vellore, India\\
  \texttt{abhinavsagar4@gmail.com} \\
}

\begin{document}

\nocite{*}

\maketitle

\begin{abstract}
In this work, we address the problem of 3D object detection from point cloud data in real time. For autonomous vehicles to work, it is very important for the perception component to detect the real world objects with both high accuracy and fast inference. We propose a novel neural network architecture along with the training and optimization details for detecting 3D objects using point cloud data. We present anchor design along with custom loss functions used in this work. A combination of spatial and channel wise attention module is used in this work. We use the Kitti 3D Bird’s Eye View dataset for benchmarking and validating our results. Our method surpasses previous state of the art in this domain both in terms of average precision and speed running at > 30 FPS. Finally, we present the ablation study to demonstrate that the performance of our network is generalizable. This makes it a feasible option to be deployed in real time applications like self driving cars.
\end{abstract}

\section{Introduction}

A lot of work has been done in 2D object detection using convolutional neural networks. The object detection algorithms can be broadly grouped into the following two types:

1. Single stage detector - Yolo \citep{redmon2016you}, SSD \citep{liu2016ssd}.

2. Two stage detector -  RCNN \citep{girshick2014rich}, Fast RCNN \citep{girshick2015fast}, Faster RCNN \citep{ren2015faster}.

The difference between the two is that in the two stage detectors, the first stage uses region proposal networks to generate regions of interest and the second stage uses these regions of interest for object classification and bounding box regression. These are proven to have achieved better accuracy than the one stage architecture but comes at a tradeoff of more computational burden and time taken. On the other hand, a single stage detector uses the input image to directly learn the class wise probability and bounding box coordinates. Thus these architectures treat the object detection as a simple regression problem and thus are faster but less accurate.

There has also been a lot of work done on 3D object detection. Some of them use a camera based approach using either monocular or stereo images. Also work has been done by fusing the depth information on RGBD images taken from the camera. The main problem with camera based approach is the low accuracy achieved. Therefore lidar data has been proven to be a better alternative achieving higher accuracy and thus safety which is a primary concern for self driving cars. The challenge with using lidar data is that it produces data in the form of point clouds which have millions of points thus increasing the computational cost and processing time. 

Point cloud data are of many types, of which the main type is 3D voxel grid. However, monocular 3D object detection is a difficult problem due to the depth information loss in 2D image planes. Recent networks have been proposed to first predict the pixel-level depth and convert the monocular image to 3D point cloud representations. These methods although achieves good performance but it introduces additional expensive computational cost for predicting high-resolution depth maps from images, making them challenging to be deployed in real time settings like self driving cars.

In this work, our approach uses only the bird's eye view for 3D object detection in real time. The context of our work is in self driving cars but can be deployed in other settings as well. To validate our work, we benchmark our results on the publicly available 3D Kitti dataset \citep{geiger2012we}. We use spatial and channel attention modules in one branch for finding out where is an informative part in the image and finding out what feature is meaningful in the image respectively. The second branch locates the 2d bounding box co-ordinates while the third branch is used to get the deviations between the predicted and actual co-ordinates. The individual features are summed to give the refined 3d bounding box co-ordinates. For the evaluation metric, we use the class wise average precision. Our work beats the previous state of the art approaches for 3D object detection while also running at greater than 30 FPS. We also further show the learning and optimization aspects along with ablation study of this approach and present how it could potentially be generalized to other real world settings.

A sample of the predicted 3D detection from the KITTI validation dataset is shown in Figure 1:

\begin{figure}[htp]
    \centering
    \includegraphics[width=12cm]{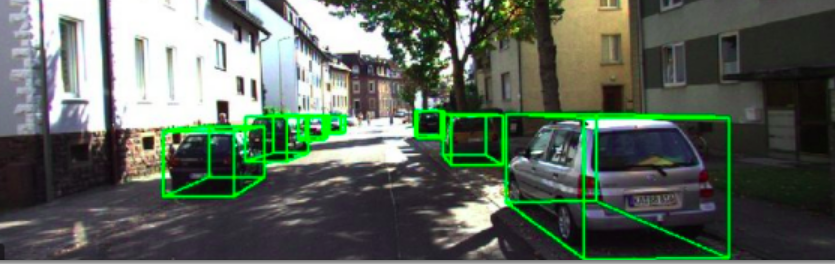}
    \caption{3D detection from the KITTI validation dataset projected onto an image}
    \label{fig8}
\end{figure}

\section{Related Work}

Recently there have been a surge of papers on 3D object detection from various kinds of data like LIDAR, stereo etc. VOTE 3D \citep{qi2019deep} uses a sliding window on a 3D  voxel grid to detect objects. The pre-trained model is fed to a SVM classifier later. VELOFCN \citep{li20173d} projects 3D point cloud data to a perspective in the front view and gets a 2D depth map. The objects are detected by running a convolutional neural network on the depth map. MV3D \citep{qi2018frustum} architecture also used a similar approach by combining the features extracted from multiple views like front view, birds eye view and camera view. These extracted features are passed through a CNN to detect 3D objects.

PointNet \citep{qi2017pointnet} proposed an end-to-end classification neural network that directly takes a point cloud as input without any preprocessing and outputs class scores. \citep{zhou2018voxelnet} subdivides the point cloud into 3D voxels and then transforms points within each voxel to a trainable feature vector that characterizes the shape information of the contained points. The representation vectors for each voxel stacks together and passes to a region proposal network to detect the objects. \citep{chen2020object} proposed and anchor free method using firing of different hotspots. \citep{ge2020afdet} used anchor free one stage network for 3d object detection. Pairwise spatial relationship of features was used for monocular 3D object detection \citep{chen2020monopair}. A learnable depth guided convolution was used to tackle monocular 3D object detection problem \citep{ding2020learning}. 

Triple attention module was used \citep{liu2020tanet} for 3d object detection from point clouds. A comprehensive study of various localization errors involved while detecting 3d objects was presented \citep{ma2021delving}. A new voting algorithm was individually proposed for improving the robustness of 3d object detector \citep{qi2020imvotenet} and \citep{xie2020mlcvnet}. \citep{zhou2020end} used an end to end learnable network using multi view feature fusion from lidar data. \citep{vora2020pointpainting} similarly used sequential fusion approach. A more generalizable method taking into account different shapes and sizes of objects present in image was proposed by \citep{zhang2021objects}. Both 3d object detection and tracking problem was tackled using a single network \citep{yin2021center}.

We summarize our main contributions as follows:

• A novel approach using spatial and channel attention mechanism to simultaneously detect and regress 3D bounding box over all the objects present in the image.

• A thorough analysis of backbone, optimization, anchors and loss function used in our network which is end to end trainable.

• Evaluation on the KITTI dataset shows we outperform
all previous state-of-the-art methods in terms of average precision while running at >30 FPS.

\section{Model}

\subsection{Dataset}

For this work, we have used the Kitti dataset \citep{geiger2012we} which contains LIDAR data taken from a sensor mounted in front of the car. Since the data contains millions of points and is of quite high resolution, processing is a challenge especially in real world situations. The task is to detect and regress a bounding box for 3D objects detected in real time. The dataset has 7481 training images and 7518 test point clouds comprising a total of labelled objects. The object detection performance is measured through average precision and IOU (Intersection over union) with threshold 0.7 for car class. The 3D object KITTI benchmark provides 3D bounding boxes for object classes including cars, vans, trucks, pedestrians and cyclists which are labelled manually in 3D point clouds on the basis of information from the camera. KITTI also provides three detection evaluation levels: easy, moderate and hard, according to the object size, occlusion state and truncation level. The minimal pixel height for easy objects is 40px, which approximately corresponds to vehicles within 28m. For moderate and hard level objects are 25px, corresponding to a minimal distance of 47m.

\subsection{Problem Definition}

Given a RGB images and the corresponding camera
parameters, our goal is to classify and localize the objects of interest in 3D space. Each object is represented
by its category, 2D bounding box $B_{2D}$, and 3D bounding
box $B_{3D}$. $B_{2D}$ is represented by its center
$c_{i}$ = $[x_{0}, y_{0}]_{2D}$ and size $[h_{0}, w_{0}]_{2D}$ in the image plane, while $B_{3D}$ is defined by its center $[x, y, z]_{3D}$, size $[h, w, l]_{3D}$ and heading angle $\gamma$ in the 3D world space.

The 3D bounding box $B_{3D}$ is the final goal of prediction.
The first task is 2D object detection in which the goal is to predict the 2D bounding box $B_{2D}$ of the object and its
class. $B_{2D} = (w_{2D}, h_{2D}, u_{b}, v_{b})$ where $(w_{2D}, h_{2D})$ indicates the size of $B_{2D}$ and $(u_{b}, v_{b})$ represents the center of $B_{2D}$ on the image plane.

\subsection{Spatial Attention Module}

The spatial attention module is used for capturing the spatial dependencies of the feature maps. The spatial attention (SA) module used in our network is defined below:

\begin{equation}
f_{{SA}}(x)=f_{sigmoid}\left({W}_{2}\left(f_{{ReLU}}\left({W}_{1}(x)\right)\right)\right)
\end{equation}

where $W_{1}$ and $W_{2}$ denotes the first and second $1 \times 1$ convolution layer respectively, $x$ denotes the input data, $f_{Sigmoid}$ denotes the sigmoid function,
$f_{ReLU}$ denotes the ReLu activation function.

The spatial attention module used in this work is shown in Figure 2:

\begin{figure}[htp]
    \centering
    \includegraphics[width=6cm]{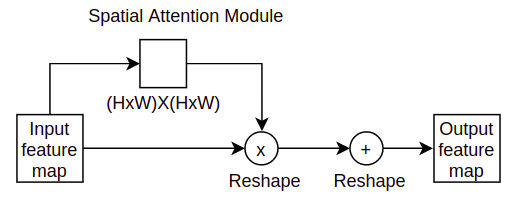}
    \caption{Illustration of our spatial attention module}.
    \label{a1}
\end{figure}

\subsection{Channel Attention Module}

The channel attention module is used for extracting high level multi-scale semantic information. The channel attention (CA) module used in our network is defined below:

\begin{equation}
f_{{CA}}(x)=f_{sigmoid }({W}_{2}(f_{{ReLU }}({W}_{1}f_{{AvgPool }}^{1}(x))))
\end{equation}

where $W_{1}$ and $W_{2}$ denotes the first and second $1 \times 1$ convolution layer, $x$ denotes the input data. $f^{1}_{AvgPool}$ denotes the global average pooling function, $f_{Sigmoid}$ denotes the Sigmoid function,
$f_{ReLU}$ denotes ReLU activation function.

The channel attention module used in this work is shown in Figure 3:

\begin{figure}[htp]
    \centering
    \includegraphics[width=6cm]{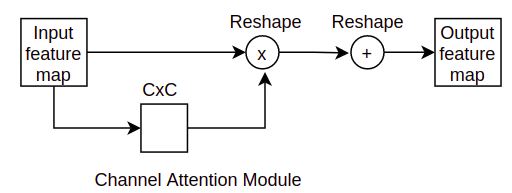}
    \caption{Illustration of our channel attention module}.
    \label{a1}
\end{figure}

\subsection{Network Architecture}

We divide the point cloud data into 3D voxel grid cells. Our CNN backbone takes as input the image in the form of voxel and outputs a feature vector. We use Resnet as backbone for our network. Residual blocks are used for locating the 2d bounding box co-ordinates which is then propagated to a Roi Align operator which is then sent to a fully connected layer. In parallel, spatial and channel attention mechanism are used for finding out where is an informative part in the image and finding out what feature is meaningful given in the image. the individual features are summed up which is in turn summed up with the first block to produce the 3d bounding box co-ordinates. In parallel, a third block uses Roi Align and fully connected layers to find out the deviations between the actual and predicted co-ordinates. Anchors are used in these deltas blocks to adjust the coordinates according to the size and shape of the object detected. This block is learnable thus improving the hyper-parameters in every iteration. The learned deviations are finally summed up with the 3d bounding box co-ordinates to give the refined 3d bounding box co-ordinates. 

The residual blocks are made up of: a fully connected layer followed by a non linearity activation function which is ReLU used in this case and a batch normalization layer. These layers are used for transforming each point in the voxel to a point wise feature vector. Element wise max-pooling layer is also used which extracts the maximum value from all the neighbouring pixel values when the filter is applied on the image. This operation is used for getting the locally aggregated features. Also a point wise concatenation operator is used which concatenates each point wise feature vector with the locally aggregated features. For our detector there are in total 7 parameters - three for the offset center coordinates, three for the offset dimensions and the last is for offset rotation angle. The network architecture is shown in Figure 4:

\begin{figure}[htp]
    \centering
    \includegraphics[width=14cm]{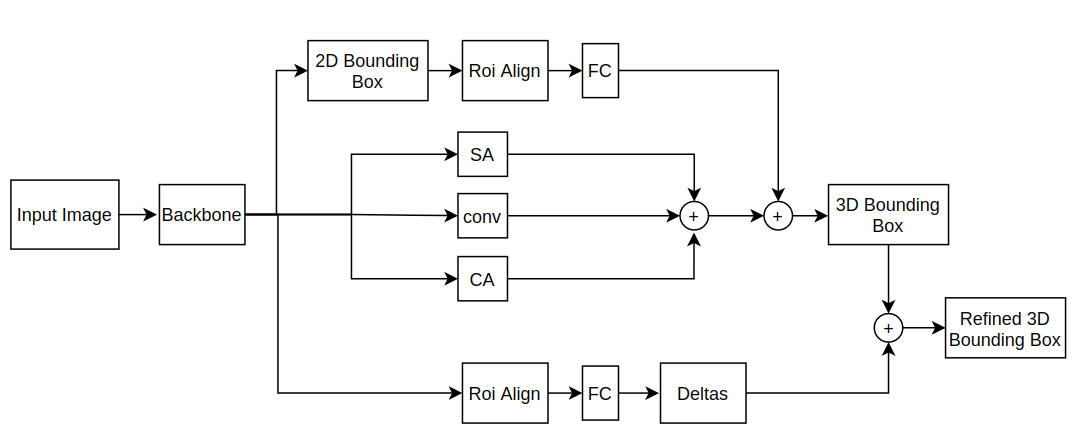}
    \caption{Illustration of our network architecture. SA denotes spatial attention module, CA denotes channel attention module, FC denotes fully connected layer and + denotes summation operator.}
    \label{fig2}
\end{figure}

\section{Experiments}

\subsection{Anchors}

Anchors are very important for efficient object detection. These are basically prior beliefs containing information of the size for the detected object, its position is the image, its pose, its orientation etc. Anchors of multiple shape, size are more stable, also helps in reducing the computational burden and time taken by the model. We have chosen two anchors for each of the classes as shown in Table 1, Table 2 and Table 3 respectively:

\begin{table}[hbt!]
  \caption{Car anchors}
  \label{c}
  \centering
  \begin{tabular}{llll}
  \toprule
    Height(m)     & Width(m) & Length(m) & Rotation(Theta)           \\
    \midrule
    1.6 & 1.6 &4 &0              \\
    1.6&1.6&1.6&90          \\
    \bottomrule
  \end{tabular}
\end{table}

\begin{table}[hbt!]
  \caption{Pedestrian anchors}
  \label{p}
  \centering
  \begin{tabular}{llll}
  \toprule
    Height(m)     & Width(m) & Length(m) & Rotation(Theta)           \\
    \midrule
    1.7 & 0.5 &0.7 &0              \\
    1.7&1.5&0.7&90          \\
    \bottomrule
  \end{tabular}
\end{table}
 
\begin{table}[hbt!]
  \caption{Cyclist anchors}
  \label{cy}
  \centering
  \begin{tabular}{llll}
  \toprule
    Height(m)     & Width(m) & Length(m) & Rotation(Theta)           \\
    \midrule
    1.6 & 0.7 &2 &0              \\
    1.6 & 0.7 &2 &90          \\
    \bottomrule
  \end{tabular}
\end{table}

\subsection{Loss Functions}

A vector $s = (x, y, z, l, h, w, \theta)$ represents 3D bounding box center coordinates, height, width, length and yaw respectively. The geometric relations between various parameters is illustrated in the equation below where $s$ represents the ground truth vector and $a$ represents the anchor vector. The localization regression between ground truth and anchors are defined using set of Equations 3-10:

\begin{equation}
\Delta x=\frac{x_{s}-x_{a}}{\sqrt{l^{2}+w^{2}}} 
\end{equation}

\begin{equation}
\Delta z_{b}=z_{s}-\frac{h_{s}}{2}-z_{a}+\frac{h_{a}}{2} 
\end{equation}

\begin{equation}
\Delta y=\frac{y_{s}-y_{a}}{\sqrt{l^{2}+w^{2}}} 
\end{equation}

\begin{equation}
\Delta z_{t}=z_{s}+\frac{h_{s}}{2}-z_{a}-\frac{h_{a}}{2} 
\end{equation}

\begin{equation}
\Delta l=\log \frac{l_{s}}{l_{a}} 
\end{equation}

\begin{equation}
\Delta w=\log \frac{w_{s}}{w_{a}} 
\end{equation}

\begin{equation}
\Delta \zeta=\left|\sin \left(\theta_{s}-\theta_{a}\right)\right| 
\end{equation}

\begin{equation}
\Delta \eta=\cos \left(\theta_{s}-\theta_{a}\right)
\end{equation}

Since the angle localization loss cannot distinguish the bounding boxes which are flipped, we use a softmax classification loss as shown for both positive and negative anchors. For the object classification, we have used focal loss as shown in Equation 11 and Equation 12 respectively:

\begin{equation}
\mathcal{L}_{pos}=-\alpha_{a}\left(1-p^{a}\right)^{\gamma} \log p^{a}
\end{equation}

\begin{equation}
\mathcal{L}_{neg}=-\alpha_{a}\left(1-p^{a}\right)^{\gamma} \log p^{a}
\end{equation}

We used Intersection Over Union (IOU) for evaluating the performance of our network. All the positive anchors have an IOU value above 0.60 while those with less than 0.45 are treated as negative anchors. We used binary cross entropy loss for detection and a variant of huber loss for regression. 

Let $i$ and $j$ denote the positive and negative anchors and let $p$ denote the sigmoid activation for the classification network. Let $pos$ represent the positive regression anchors and $neg$ the negative regression anchors. The individual loss terms can be denoted using set of Equations 13-15.

\begin{equation}
L_{1} = \frac{1}{N} \sum_{i} L_{p o s}\left(p_{i}^{p o s}, 1\right)
\end{equation}

\begin{equation}
L_{2} = \frac{1}{N} \sum_{j} L_{n e g}\left(p_{j}^{n e g}, 0\right)
\end{equation}

\begin{equation}
L_{3} = \frac{1}{N} \sum_{k}\left(L_{r}\left(l, l^{*}\right)+L\left(h, h^{*}\right)+L_{c}\left(w, w^{*}\right)\right)
\end{equation}

The overall loss function is shown in Equation 16:

\begin{equation}
L_{total} = \alpha L_{1} + \beta L_{2} + \gamma L_{3} 
\end{equation}

Here $\alpha$, $\beta$ and $\gamma$ are the hyper-parameters with values set as 0.5, 0.5 and 1.0 respectively.

\subsection{Evaluation Metrics}

We use the Average Precision with 40 recall positions ($AP_{40}$) under three difficult settings (easy, moderate, and hard) for those tasks. We present the
performances of the Car, Pedestrian and Cyclist categories as reference. The default IoU threshold values are 0.7, 0.5, 0.5 for these three categories respectively. Each manually annotated object is divided into easy, moderate, and hard level according to the occlusion, truncation, and box height. The metrics used extensively in the literature are Average precisions (AP)
on the car class for bird’s-eye-view (BEV) and 3D boxes
with 0.5/0.7 IoU thresholds. We present both $AP_{11}$ and
$AP_{40}$ results to make comprehensive comparisons as has been studied in literature.

\subsection{Implementation Details}

We train our model on a GTX 1080Ti GPU with a batch size of 16 for 100 epochs. We use Adam optimizer with an initial learning rate of 0.001, and decay it by ten times at every 100 epochs.
The weight decay is set to 0.0001. We use Non-Maximum Suppression (NMS) on center detection results. We use 3D
bounding boxes score of center detection as the
confidence of predicted results. We discard predictions with confidence value less than 0.1. All input images are
padded to the same size of $384 \times 1280$. The prediction
head of the backbone consists of one $3 \times 3 \times 256$
conv layer, BatchNorm, ReLU, and $1 \times 1 \times op$
conv layer where $op$ is the output size. 

\section{Results}

We report our results of the Car category on KITTI test set as shown in Table 4. Overall, our method achieves superior results over previous methods. Compared with the methods with extra data, our network still get comparable performances, which further proves the effectiveness of our model. Our method is also much faster than most existing methods, allowing for real-time inference which is important in the context of autonomous driving.

\begin{table}[hbt!]
\tiny
  \caption{Quantitative results for Car on KITTI test sets, evaluated by AP3D. “Extra” lists the required extra information for each method. We divide existing methods into two groups considering whether they utilize extra information and sort them according to their performance on the moderate level of the test set within each group. The three sets of Easy, Mod and Hard denotes Val $AP_{11}$, Val $AP_{40}$ Test $AP_{40}$ respectively.}
  \label{sample-table3}
  \centering
  \begin{tabular}{llllllllllll}
  \toprule
    Method &Extra &Time(ms) &$Easy_{1}$ &$Mod_{1}$ &$Hard_{1}$ &$Easy_{2}$ &$Mod_{2}$ &$Hard_{2}$ &$Easy_{3}$ &$Mod_{3}$ &$Hard_{3}$\\
    \midrule
MonoPSR &depth, LiDAR &120 &12.75 &11.48 &8.59 &- &- &- &10.76& 7.25 &5.85\\
UR3D &depth &120 &28.05 &18.76 &16.55 &23.24 &13.35 &10.15 &15.58 &8.61 &6.00\\
AM3D &depth &- &32.23 &21.09 &17.26 &28.31 &15.76 &12.24 &16.50& 10.74 &9.52\\
PatchNet &depth &- &35.10 &22.00 &19.60 &31.60 &16.80 &13.80& 15.68 &11.12 &10.17\\
DA-3Ddet &depth, LiDAR &- &33.40 &24.00 &19.90 &- &- &- &16.80& 11.50 &8.90\\
D4LCN &depth &- &26.97 &21.71 &18.22 &22.32 &16.20 &12.30 &16.65& 11.72 &9.51\\
Kinem3D &multi-frames &120 &- &- &- &19.76 &14.10 &10.47 &19.07& 12.72 &9.17\\
FQNet &- &- &5.98 &5.50 &4.75 &- &- &- &2.77 &1.51 &1.01\\
MonoGRNet &- &60 &13.88 &10.19 &7.62 &- &- &- &9.61 &5.74 &4.25\\
MonoDIS &- &100 &18.05 &14.98 &13.42 &- &- &- &10.37 &7.94 &6.40\\
M3D-RPN &- &160 &20.27 &17.06 &15.21 &14.53 &11.07 &8.65 &14.76& 9.71 &7.42\\
MonoPair &- &57 &- &- &- &16.28 &12.30 &10.42 &13.04 &9.99 &8.65\\
RTM3D &- &55 &20.77 &16.86 &16.63 &- &- &- &14.41 &10.34 &8.77\\
Movi3D &- &45 &- &- &- &14.28 &11.13 &9.68 &15.19 &10.90 &9.26\\
\cite{zhang2021objects} &- &35 &28.17 &21.92 &19.07 &23.64 &17.51 &14.83 &19.94& 13.89 &12.07\\
AA3DNet &- &\textbf{26} &\textbf{30.22} &\textbf{22.54} &18.38 &\textbf{24.01} &\textbf{17.81} &14.31 &\textbf{21.62}& \textbf{14.90} &11.82\\
    \bottomrule
  \end{tabular}
\end{table}

We present our model’s performance on the KITTI validation set in Table 5. Our approach shows better performance consistency between the validation set and test set. This indicates that our method has better generalization ability, which is important in autonomous driving.

\begin{table}[hbt!]
\tiny
  \caption{Performance of the Car category on the KITTI validation set. Methods are ranked by moderate setting (same as KITTI leaderboard). We highlight the best results in bold. The four sets of Easy, Mod and Hard denotes $3D_{IOU}$=0.7, $BEV_{IOU}$=0.7, $3D_{IOU}$=0.5 and $BEV_{IOU}$=0.5 respectively.}
  \label{sample-table4}
  \centering
  \begin{tabular}{lllllllllllll}
  \toprule
Method &$Easy_{1}$ &$Mod_{1}$ &$Hard{1}$ &$Easy_{2}$ &$Mod_{2}$ &$Hard_{2}$ &$Easy_{3}$ &$Mod_{3}$ &$Hard_{3}$ &$Easy_{4}$ &$Mod_{4}$ &$Hard_{4}$\\
    \midrule
CenterNet &0.60 &0.66 &0.77 &3.46 &3.31 &3.21 &20.00 &17.50& 15.57 &34.36 &27.91 &24.65\\
MonoGRNet &11.90 &7.56 &5.76 &19.72 &12.81 &10.15 &47.59 &32.28 &25.50 &48.53 &35.94 &28.59\\
MonoDIS &11.06 &7.60 &6.37 &18.45 &12.58 &10.66 &- &- &- &-\\
M3D-RPN &14.53 &11.07 &8.65 &20.85 &15.62 &11.88 &48.53 &35.94& 28.59 &53.35 &39.60 &31.76\\
MonoPair &16.28 &12.30 &10.42 &24.12 &18.17 &15.76 &55.38 &42.39& 37.99 &61.06 &47.63 &41.92\\
\citep{ma2021delving} &17.45 &13.66 &11.68 &24.97 &19.33 &17.01 &55.41 &43.42 &37.81 &60.73 &46.87 &41.89\\
AA3DNet &\textbf{18.06} &\textbf{14.27} &11.51 &\textbf{25.68} &\textbf{19.83} &16.64 &\textbf{57.24} &\textbf{44.90} &37.15 &\textbf{62.18} &\textbf{47.55} &41.24\\
    \bottomrule
  \end{tabular}
\end{table}

Our results are considerably better than the previous state of the art approaches.

\subsection{Average Precision}

The ideal value of precision and recall is 1. Since it is not possible to get perfect values, the closer the metrics ie precision and recall is to 1, the better our model is performing, It’s often seen that there is a tradeoff between precision and recall ie if we are optimizing for precision, recall value gets less and if we are trying to improve recall, precision value becomes less. So our task is to balance both and note that threshold point. Average precision is the average value of precision for the sampled points at various recall threshold values. The precision -  recall curve for 3D object detection for the 3 classes i.e. cars, pedestrians and cyclists for all the three categories i.e. easy, moderate and hard are shown in Figure 5. The closer the curve is to (1,1), the higher performance of the model is.

\begin{figure}[htp]
    \centering
    \includegraphics[width=10cm]{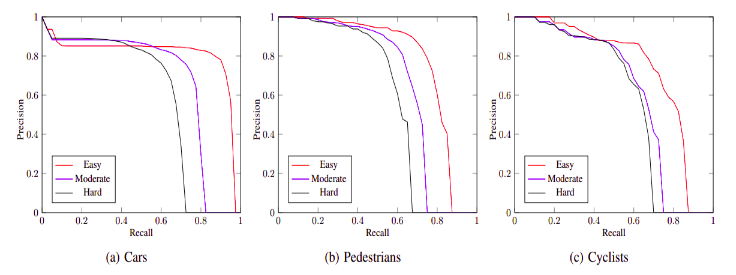}
    \caption{Precision-recall curve for 3D detection in a) Cars b) Pedestrian c) Cyclists.}
    \label{fig3}
\end{figure}

Finally we present the results for 3D object detection results on KITTI validation set in Figure 6. The ground truth bounding boxes are shown in blue and the predicted bounding boxes are shown in orange.

\begin{figure}[htp]
    \centering
    \includegraphics[width=10cm]{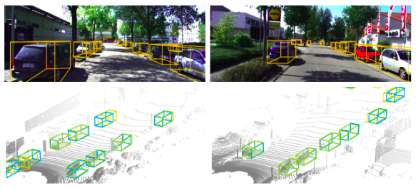}
    \caption{Predicted 3D bounding boxes are drawn in orange, while ground truths are in blue.
}
    \label{fig4}
\end{figure}

Note that our model is based only on LiDAR data. For better visualization the 3D bounding boxes are projected on to the bird’s eye view and the images.

\subsection{Ablation Study}

The compared results with different backbones on Average Precision metric is shown in Table 6:

\begin{table}[hbt!]
  \caption{Ablation study of different backbone networks on $AP_{3D}$ (IoU=0.3).}
  \label{sample-table1}
  \centering
  \begin{tabular}{llll}
  \toprule
    Backbone Network & Easy &Moderate &Hard         \\
    \midrule
    VGG16 &53.68 &41.45 &34.08\\
InceptionV3 &54.32 &41.60 &34.66\\
DenseNet169 &54.26 &40.04 &35.06\\
ResNet50 &\textbf{56.16} &\textbf{42.61} &\textbf{35.36}\\
    \bottomrule
  \end{tabular}
\end{table}

The best results are achieved using ResNet50 as the backbone on our network.

A study of with and without using channel and spatial attention module on Average Precision metric is shown in Table 7:

\begin{table}[h]
  \caption{Ablation study using variations of spatial and channel attention modules on $AP_{3D}$ (IoU=0.3).}
  \label{sample-table2}
  \centering
  \begin{tabular}{llll}
  \toprule
Attention Module &Easy &Moderate &Hard \\
   \midrule
No attention &53.59 &40.06 &32.18\\  
Only SA &55.05 &42.06 &34.58\\
Only CA &55.51 &40.49 &34.46\\
Both &\textbf{56.16} &\textbf{42.61} &\textbf{35.36}\\
    \bottomrule
  \end{tabular}
\end{table}

The best results are achieved using both spatial and channel attention modules in our network.

A study of using individual loss function terms used while training our network on Average Precision metric is shown in Table 8:

\begin{table}[h]
  \caption{Ablation study using individual loss function terms on $AP_{3D}$ (IoU=0.3).}
  \label{sample-table2}
  \centering
  \begin{tabular}{llllll}
  \toprule
$L_{1}$ &$L_{2}$ &$L_{3}$ &Easy &Moderate &Hard\\
\midrule
$\times$ &$\times$ &$\checkmark$ &44.50 &32.33 &29.10\\
$\checkmark$ &$\checkmark$ &$\times$ &52.72 &40.59 &33.71\\
$\checkmark$ &$\checkmark$ &$\checkmark$ &\textbf{56.16} &\textbf{42.61} &\textbf{35.36}\\
    \bottomrule
  \end{tabular}
\end{table}

The best results are achieved using all the loss functions ie $L_{1}$, $L_{2}$ and $L_{3}$ combined.

\section{Conclusions}

In this paper, we proposed a real time 3D object detection network using spatial and channel attention mechanism using LIDAR point cloud data. For making efficient computation, our architecture uses a single stage type neural network with bird's view representation. We evaluate our network on the KITTI benchmark dataset and show that our approach outperforms previous state of the art methids. As for the evaluation metric, we chose class wise average precision. The model runs at faster than 30 FPS and hence can be used in autonomous driving applications where safety is a major challenge. In the future, we would be interested in studying attention mechanism in the context of 3D semantic segmentation.

\subsubsection*{Acknowledgments}

We would like to thank Nvidia for providing the GPUs.

\bibliography{neurips_2019}

\begin{thebibliography}{44}
\providecommand{\natexlab}[1]{#1}
\providecommand{\url}[1]{\texttt{#1}}
\expandafter\ifx\csname urlstyle\endcsname\relax
  \providecommand{\doi}[1]{doi: #1}\else
  \providecommand{\doi}{doi: \begingroup \urlstyle{rm}\Url}\fi

\bibitem[Chen et~al.(2020{\natexlab{a}})Chen, Sun, Wang, Jia, and
  Yuille]{chen2020object}
Q.~Chen, L.~Sun, Z.~Wang, K.~Jia, and A.~Yuille.
\newblock Object as hotspots: An anchor-free 3d object detection approach via
  firing of hotspots.
\newblock In \emph{European Conference on Computer Vision}, pages 68--84.
  Springer, 2020{\natexlab{a}}.

\bibitem[Chen et~al.(2016)Chen, Kundu, Zhang, Ma, Fidler, and
  Urtasun]{chen2016monocular}
X.~Chen, K.~Kundu, Z.~Zhang, H.~Ma, S.~Fidler, and R.~Urtasun.
\newblock Monocular 3d object detection for autonomous driving.
\newblock In \emph{Proceedings of the IEEE Conference on Computer Vision and
  Pattern Recognition}, pages 2147--2156, 2016.

\bibitem[Chen et~al.(2017)Chen, Ma, Wan, Li, and Xia]{chen2017multi}
X.~Chen, H.~Ma, J.~Wan, B.~Li, and T.~Xia.
\newblock Multi-view 3d object detection network for autonomous driving.
\newblock In \emph{Proceedings of the IEEE Conference on Computer Vision and
  Pattern Recognition}, pages 1907--1915, 2017.

\bibitem[Chen et~al.(2020{\natexlab{b}})Chen, Liu, Shen, and Jia]{chen2020dsgn}
Y.~Chen, S.~Liu, X.~Shen, and J.~Jia.
\newblock Dsgn: Deep stereo geometry network for 3d object detection.
\newblock In \emph{Proceedings of the IEEE/CVF Conference on Computer Vision
  and Pattern Recognition}, pages 12536--12545, 2020{\natexlab{b}}.

\bibitem[Chen et~al.(2020{\natexlab{c}})Chen, Tai, Sun, and
  Li]{chen2020monopair}
Y.~Chen, L.~Tai, K.~Sun, and M.~Li.
\newblock Monopair: Monocular 3d object detection using pairwise spatial
  relationships.
\newblock In \emph{Proceedings of the IEEE/CVF Conference on Computer Vision
  and Pattern Recognition}, pages 12093--12102, 2020{\natexlab{c}}.

\bibitem[Dai et~al.(2016)Dai, Li, He, and Sun]{dai2016r}
J.~Dai, Y.~Li, K.~He, and J.~Sun.
\newblock R-fcn: Object detection via region-based fully convolutional
  networks.
\newblock In \emph{Advances in neural information processing systems}, pages
  379--387, 2016.

\bibitem[Ding et~al.(2020)Ding, Huo, Yi, Wang, Shi, Lu, and
  Luo]{ding2020learning}
M.~Ding, Y.~Huo, H.~Yi, Z.~Wang, J.~Shi, Z.~Lu, and P.~Luo.
\newblock Learning depth-guided convolutions for monocular 3d object detection.
\newblock In \emph{Proceedings of the IEEE/CVF Conference on Computer Vision
  and Pattern Recognition Workshops}, pages 1000--1001, 2020.

\bibitem[Engelcke et~al.(2017)Engelcke, Rao, Wang, Tong, and
  Posner]{engelcke2017vote3deep}
M.~Engelcke, D.~Rao, D.~Z. Wang, C.~H. Tong, and I.~Posner.
\newblock Vote3deep: Fast object detection in 3d point clouds using efficient
  convolutional neural networks.
\newblock In \emph{2017 IEEE International Conference on Robotics and
  Automation (ICRA)}, pages 1355--1361. IEEE, 2017.

\bibitem[Ge et~al.(2020)Ge, Ding, Hu, Wang, Chen, Huang, and Li]{ge2020afdet}
R.~Ge, Z.~Ding, Y.~Hu, Y.~Wang, S.~Chen, L.~Huang, and Y.~Li.
\newblock Afdet: Anchor free one stage 3d object detection.
\newblock \emph{arXiv preprint arXiv:2006.12671}, 2020.

\bibitem[Geiger et~al.(2012)Geiger, Lenz, and Urtasun]{geiger2012we}
A.~Geiger, P.~Lenz, and R.~Urtasun.
\newblock Are we ready for autonomous driving? the kitti vision benchmark
  suite.
\newblock In \emph{2012 IEEE Conference on Computer Vision and Pattern
  Recognition}, pages 3354--3361. IEEE, 2012.

\bibitem[Girshick(2015)]{girshick2015fast}
R.~Girshick.
\newblock Fast r-cnn.
\newblock In \emph{Proceedings of the IEEE international conference on computer
  vision}, pages 1440--1448, 2015.

\bibitem[Girshick et~al.(2014)Girshick, Donahue, Darrell, and
  Malik]{girshick2014rich}
R.~Girshick, J.~Donahue, T.~Darrell, and J.~Malik.
\newblock Rich feature hierarchies for accurate object detection and semantic
  segmentation.
\newblock In \emph{Proceedings of the IEEE conference on computer vision and
  pattern recognition}, pages 580--587, 2014.

\bibitem[He et~al.(2020)He, Zeng, Huang, Hua, and Zhang]{he2020structure}
C.~He, H.~Zeng, J.~Huang, X.-S. Hua, and L.~Zhang.
\newblock Structure aware single-stage 3d object detection from point cloud.
\newblock In \emph{Proceedings of the IEEE/CVF Conference on Computer Vision
  and Pattern Recognition}, pages 11873--11882, 2020.

\bibitem[Huang et~al.(2020)Huang, Liu, Chen, and Bai]{huang2020epnet}
T.~Huang, Z.~Liu, X.~Chen, and X.~Bai.
\newblock Epnet: Enhancing point features with image semantics for 3d object
  detection.
\newblock In \emph{European Conference on Computer Vision}, pages 35--52.
  Springer, 2020.

\bibitem[Ku et~al.(2018)Ku, Mozifian, Lee, Harakeh, and Waslander]{ku2018joint}
J.~Ku, M.~Mozifian, J.~Lee, A.~Harakeh, and S.~L. Waslander.
\newblock Joint 3d proposal generation and object detection from view
  aggregation.
\newblock In \emph{2018 IEEE/RSJ International Conference on Intelligent Robots
  and Systems (IROS)}, pages 1--8. IEEE, 2018.

\bibitem[Li et~al.(2016)Li, Zhang, and Xia]{li2016vehicle}
B.~Li, T.~Zhang, and T.~Xia.
\newblock Vehicle detection from 3d lidar using fully convolutional network.
\newblock \emph{arXiv preprint arXiv:1608.07916}, 2016.

\bibitem[Liang et~al.(2019)Liang, Yang, Chen, Hu, and Urtasun]{liang2019multi}
M.~Liang, B.~Yang, Y.~Chen, R.~Hu, and R.~Urtasun.
\newblock Multi-task multi-sensor fusion for 3d object detection.
\newblock In \emph{Proceedings of the IEEE Conference on Computer Vision and
  Pattern Recognition}, pages 7345--7353, 2019.

\bibitem[Lin et~al.(2017{\natexlab{a}})Lin, Doll{\'a}r, Girshick, He,
  Hariharan, and Belongie]{lin2017feature}
T.-Y. Lin, P.~Doll{\'a}r, R.~Girshick, K.~He, B.~Hariharan, and S.~Belongie.
\newblock Feature pyramid networks for object detection.
\newblock In \emph{Proceedings of the IEEE conference on computer vision and
  pattern recognition}, pages 2117--2125, 2017{\natexlab{a}}.

\bibitem[Lin et~al.(2017{\natexlab{b}})Lin, Goyal, Girshick, He, and
  Doll{\'a}r]{lin2017focal}
T.-Y. Lin, P.~Goyal, R.~Girshick, K.~He, and P.~Doll{\'a}r.
\newblock Focal loss for dense object detection.
\newblock In \emph{Proceedings of the IEEE international conference on computer
  vision}, pages 2980--2988, 2017{\natexlab{b}}.

\bibitem[Liu et~al.(2016)Liu, Anguelov, Erhan, Szegedy, Reed, Fu, and
  Berg]{liu2016ssd}
W.~Liu, D.~Anguelov, D.~Erhan, C.~Szegedy, S.~Reed, C.-Y. Fu, and A.~C. Berg.
\newblock Ssd: Single shot multibox detector.
\newblock In \emph{European conference on computer vision}, pages 21--37.
  Springer, 2016.

\bibitem[Liu et~al.(2020)Liu, Zhao, Huang, Hu, Zhou, and Bai]{liu2020tanet}
Z.~Liu, X.~Zhao, T.~Huang, R.~Hu, Y.~Zhou, and X.~Bai.
\newblock Tanet: Robust 3d object detection from point clouds with triple
  attention.
\newblock In \emph{Proceedings of the AAAI Conference on Artificial
  Intelligence}, volume~34, pages 11677--11684, 2020.

\bibitem[Ma et~al.(2021)Ma, Zhang, Xu, Zhou, Yi, Li, and Ouyang]{ma2021delving}
X.~Ma, Y.~Zhang, D.~Xu, D.~Zhou, S.~Yi, H.~Li, and W.~Ouyang.
\newblock Delving into localization errors for monocular 3d object detection.
\newblock In \emph{Proceedings of the IEEE/CVF Conference on Computer Vision
  and Pattern Recognition}, pages 4721--4730, 2021.

\bibitem[Peng et~al.(2021)Peng, Liu, Yu, Yan, Deng, and Cai]{peng2021lidar}
L.~Peng, F.~Liu, Z.~Yu, S.~Yan, D.~Deng, and D.~Cai.
\newblock Lidar point cloud guided monocular 3d object detection.
\newblock \emph{arXiv preprint arXiv:2104.09035}, 2021.

\bibitem[Qi et~al.(2017)Qi, Su, Mo, and Guibas]{qi2017pointnet}
C.~R. Qi, H.~Su, K.~Mo, and L.~J. Guibas.
\newblock Pointnet: Deep learning on point sets for 3d classification and
  segmentation.
\newblock In \emph{Proceedings of the IEEE conference on computer vision and
  pattern recognition}, pages 652--660, 2017.

\bibitem[Qi et~al.(2018)Qi, Liu, Wu, Su, and Guibas]{qi2018frustum}
C.~R. Qi, W.~Liu, C.~Wu, H.~Su, and L.~J. Guibas.
\newblock Frustum pointnets for 3d object detection from rgb-d data.
\newblock In \emph{Proceedings of the IEEE conference on computer vision and
  pattern recognition}, pages 918--927, 2018.

\bibitem[Qi et~al.(2019)Qi, Litany, He, and Guibas]{qi2019deep}
C.~R. Qi, O.~Litany, K.~He, and L.~J. Guibas.
\newblock Deep hough voting for 3d object detection in point clouds.
\newblock In \emph{Proceedings of the IEEE International Conference on Computer
  Vision}, pages 9277--9286, 2019.

\bibitem[Qi et~al.(2020)Qi, Chen, Litany, and Guibas]{qi2020imvotenet}
C.~R. Qi, X.~Chen, O.~Litany, and L.~J. Guibas.
\newblock Imvotenet: Boosting 3d object detection in point clouds with image
  votes.
\newblock In \emph{Proceedings of the IEEE/CVF conference on computer vision
  and pattern recognition}, pages 4404--4413, 2020.

\bibitem[Qian et~al.(2020)Qian, Garg, Wang, You, Belongie, Hariharan, Campbell,
  Weinberger, and Chao]{qian2020end}
R.~Qian, D.~Garg, Y.~Wang, Y.~You, S.~Belongie, B.~Hariharan, M.~Campbell,
  K.~Q. Weinberger, and W.-L. Chao.
\newblock End-to-end pseudo-lidar for image-based 3d object detection.
\newblock In \emph{Proceedings of the IEEE/CVF Conference on Computer Vision
  and Pattern Recognition}, pages 5881--5890, 2020.

\bibitem[Qin et~al.(2021)Qin, Wang, and Lu]{qin2021monogrnet}
Z.~Qin, J.~Wang, and Y.~Lu.
\newblock Monogrnet: A general framework for monocular 3d object detection.
\newblock \emph{IEEE Transactions on Pattern Analysis and Machine
  Intelligence}, 2021.

\bibitem[Redmon et~al.(2016)Redmon, Divvala, Girshick, and
  Farhadi]{redmon2016you}
J.~Redmon, S.~Divvala, R.~Girshick, and A.~Farhadi.
\newblock You only look once: Unified, real-time object detection.
\newblock In \emph{Proceedings of the IEEE conference on computer vision and
  pattern recognition}, pages 779--788, 2016.

\bibitem[Ren et~al.(2015)Ren, He, Girshick, and Sun]{ren2015faster}
S.~Ren, K.~He, R.~Girshick, and J.~Sun.
\newblock Faster r-cnn: Towards real-time object detection with region proposal
  networks.
\newblock In \emph{Advances in neural information processing systems}, pages
  91--99, 2015.

\bibitem[Sagar(2021)]{sagar2021dmsanet}
A.~Sagar.
\newblock Dmsanet: Dual multi scale attention network.
\newblock \emph{arXiv preprint arXiv:2106.08382}, 2021.

\bibitem[Sagar and Soundrapandiyan(2020)]{sagar2020semantic}
A.~Sagar and R.~Soundrapandiyan.
\newblock Semantic segmentation with multi scale spatial attention for self
  driving cars.
\newblock \emph{arXiv preprint arXiv:2007.12685}, 2020.

\bibitem[Shi et~al.(2019)Shi, Wang, and Li]{shi2019pointrcnn}
S.~Shi, X.~Wang, and H.~Li.
\newblock Pointrcnn: 3d object proposal generation and detection from point
  cloud.
\newblock In \emph{Proceedings of the IEEE Conference on Computer Vision and
  Pattern Recognition}, pages 770--779, 2019.

\bibitem[Shi et~al.(2020{\natexlab{a}})Shi, Guo, Jiang, Wang, Shi, Wang, and
  Li]{shi2020pv}
S.~Shi, C.~Guo, L.~Jiang, Z.~Wang, J.~Shi, X.~Wang, and H.~Li.
\newblock Pv-rcnn: Point-voxel feature set abstraction for 3d object detection.
\newblock In \emph{Proceedings of the IEEE/CVF Conference on Computer Vision
  and Pattern Recognition}, pages 10529--10538, 2020{\natexlab{a}}.

\bibitem[Shi et~al.(2020{\natexlab{b}})Shi, Wang, Shi, Wang, and
  Li]{shi2020points}
S.~Shi, Z.~Wang, J.~Shi, X.~Wang, and H.~Li.
\newblock From points to parts: 3d object detection from point cloud with
  part-aware and part-aggregation network.
\newblock \emph{IEEE transactions on pattern analysis and machine
  intelligence}, 2020{\natexlab{b}}.

\bibitem[Vora et~al.(2020)Vora, Lang, Helou, and
  Beijbom]{vora2020pointpainting}
S.~Vora, A.~H. Lang, B.~Helou, and O.~Beijbom.
\newblock Pointpainting: Sequential fusion for 3d object detection.
\newblock In \emph{Proceedings of the IEEE/CVF conference on computer vision
  and pattern recognition}, pages 4604--4612, 2020.

\bibitem[Xie et~al.(2020)Xie, Lai, Wu, Wang, Zhang, Xu, and
  Wang]{xie2020mlcvnet}
Q.~Xie, Y.-K. Lai, J.~Wu, Z.~Wang, Y.~Zhang, K.~Xu, and J.~Wang.
\newblock Mlcvnet: Multi-level context votenet for 3d object detection.
\newblock In \emph{Proceedings of the IEEE/CVF conference on computer vision
  and pattern recognition}, pages 10447--10456, 2020.

\bibitem[Yang et~al.(2018)Yang, Luo, and Urtasun]{yang2018pixor}
B.~Yang, W.~Luo, and R.~Urtasun.
\newblock Pixor: Real-time 3d object detection from point clouds.
\newblock In \emph{Proceedings of the IEEE conference on Computer Vision and
  Pattern Recognition}, pages 7652--7660, 2018.

\bibitem[Ye et~al.(2020)Ye, Xu, and Cao]{ye2020hvnet}
M.~Ye, S.~Xu, and T.~Cao.
\newblock Hvnet: Hybrid voxel network for lidar based 3d object detection.
\newblock In \emph{Proceedings of the IEEE/CVF conference on computer vision
  and pattern recognition}, pages 1631--1640, 2020.

\bibitem[Yin et~al.(2021)Yin, Zhou, and Krahenbuhl]{yin2021center}
T.~Yin, X.~Zhou, and P.~Krahenbuhl.
\newblock Center-based 3d object detection and tracking.
\newblock In \emph{Proceedings of the IEEE/CVF Conference on Computer Vision
  and Pattern Recognition}, pages 11784--11793, 2021.

\bibitem[Zhang et~al.(2021)Zhang, Lu, and Zhou]{zhang2021objects}
Y.~Zhang, J.~Lu, and J.~Zhou.
\newblock Objects are different: Flexible monocular 3d object detection.
\newblock In \emph{Proceedings of the IEEE/CVF Conference on Computer Vision
  and Pattern Recognition}, pages 3289--3298, 2021.

\bibitem[Zhou and Tuzel(2018)]{zhou2018voxelnet}
Y.~Zhou and O.~Tuzel.
\newblock Voxelnet: End-to-end learning for point cloud based 3d object
  detection.
\newblock In \emph{Proceedings of the IEEE Conference on Computer Vision and
  Pattern Recognition}, pages 4490--4499, 2018.

\bibitem[Zhou et~al.(2020)Zhou, Sun, Zhang, Anguelov, Gao, Ouyang, Guo, Ngiam,
  and Vasudevan]{zhou2020end}
Y.~Zhou, P.~Sun, Y.~Zhang, D.~Anguelov, J.~Gao, T.~Ouyang, J.~Guo, J.~Ngiam,
  and V.~Vasudevan.
\newblock End-to-end multi-view fusion for 3d object detection in lidar point
  clouds.
\newblock In \emph{Conference on Robot Learning}, pages 923--932. PMLR, 2020.

\end{thebibliography}

\end{document}